\documentclass[10pt,twocolumn,letterpaper]{article}
\usepackage{framed,multirow}
\usepackage{amssymb}
\usepackage{latexsym}
\usepackage{subfigure}
\usepackage{amsmath}
\usepackage{chngcntr}
\usepackage{cite}
\newcommand{\tabincell}[2]{\begin{tabular}{@{}#1@{}}#2\end{tabular}}
\usepackage{url}
\usepackage{xcolor}
\definecolor{newcolor}{rgb}{.8,.349,.1}
\usepackage[colorinlistoftodos]{todonotes}  
\usepackage{authblk}
\usepackage{geometry}
\usepackage{natbib}
\usepackage{url}
\usepackage{xcolor}
\definecolor{newcolor}{rgb}{.8,.349,.1}
\usepackage{amsmath}
\usepackage{multirow}

\begin{document}

\title{SID: Incremental Learning for Anchor-Free Object Detection via Selective and Inter-Related Distillation}
\author{Can Peng}
\author{Kun Zhao}
\author{Sam Maksoud}
\author{Meng Li}
\author{Brian C. Lovell}
\affil{School of ITEE, The University of Queensland, Brisbane, QLD, Australia}
\date{}
\maketitle

\begin{abstract}
Incremental learning requires a model to continually learn new tasks from streaming data.
However, traditional fine-tuning of a well-trained deep neural network on a new task will dramatically degrade performance on the old task --- a problem known as catastrophic forgetting.
In this paper, we address this issue in the context of anchor-free object detection, which is a new trend in computer vision as it is simple, fast, and flexible.
Simply adapting current incremental learning strategies fails on these anchor-free detectors due to lack of consideration of their specific model structures.
To deal with the challenges of incremental learning on anchor-free object detectors, we propose a novel incremental learning paradigm called Selective and Inter-related Distillation (SID).
In addition, a novel evaluation metric is proposed to better assess the performance of detectors under incremental learning conditions.
By selective distilling at the proper locations and further transferring additional instance relation knowledge, our method demonstrates significant advantages on the benchmark datasets PASCAL VOC and COCO.
\end{abstract}

\section{Introduction}
\label{Introduction}
Due to the rapid development of deep neural networks (DNNs), significant advances have been achieved on many computer vision applications.
The traditional method of supervised training of a DNN requires access to labeled data where the number of classes for the task is predefined.
However, in many real-life applications, the model needs to gradually learn new classes from streaming data.
For example, in microbiology, an object detector may be trained to detect several known types of bacteria.
However, at a later date, the pathologist might want the detector to be able to detect new types of bacteria as well as all previous types of bacteria.
Unfortunately, at this time the original labelled training data might be inaccessible due to privacy, storage or license problems.
Directly fine-tuning the model in a scenario where only new class data is available will cause a dramatic drop in performance on the old classes --- this is a well-known problem called catastrophic forgetting \citep{goodfellow2013empirical, mccloskey1989catastrophic}.
To tackle this problem, incremental learning techniques have been widely explored in image classification \citep{li2017learning, rebuffi2017icarl} and also in anchor-based object detection \citep{shmelkov2017incremental}.

Modern Anchor-free fully convolutional object detectors were proposed to achieve superior performance compared to older anchor-based counterparts.
Compared to anchor-based detectors, anchor-free detectors are more efficient and avoid complex hyper-parameter tuning related to anchor boxes.
Also, object detection seems to be the only prediction task that deviates from the fully convolutional per-pixel prediction framework due to the need for clumsy anchor boxes.
Anchor-free detectors provide a research direction for designing a single model to perform multiple prediction tasks.
Therefore, these anchor-free methods have become the new trend in object detection.
Naturally, this new trend creates new challenges for incremental object detection.
Exploring the fully convolutional anchor-free incremental object detection is an important step towards developing a universal incremental model that can perform multiple incremental prediction tasks.

\begin{figure}[tbp]
	\centering
	\includegraphics[width=8cm, keepaspectratio]{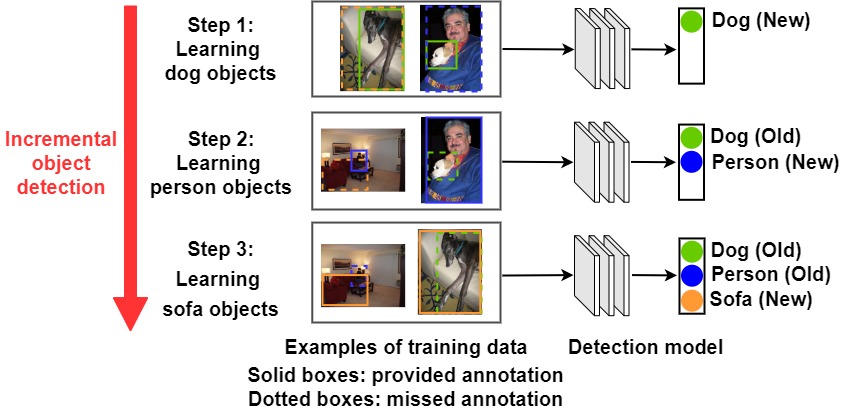}
	\caption{An example of incremental object detection.
		The challenge is to retrain the model to maintain the detection of the old classes whilst learning the new classes.
		During each incremental step, only new class data and annotations for new class objects are provided.}
	\label{fig:incremental learning}
\end{figure}

Figure \ref{fig:incremental learning} illustrates an example of incremental object detection.
Given an object detection model that is trained using images from certain classes, incremental object detection is the task of retraining the model to maintain the detection performance on the previously learned classes whilst learning new classes.
In multi-step incremental detection, all categories trained during any previous steps are regarded as old classes.
The original model is regarded as the source model (old model) and the retrained model is regarded as the target model (new model).
Knowledge distillation \citep{hinton2015distilling} is a popular strategy for incremental learning and has shown superior performance in both classification and detection tasks.
LwF \citep{li2017learning} was the first to explore incremental learning by knowledge distillation.
It uses the source model output itself as the ground truth to guide the target model to follow the behavior of the source model on the old classes.
\citet{shmelkov2017incremental} adapted LwF for incremental detection by distilling the classification and bounding box regression outputs for a two-stage anchor-based detector (Fast RCNN).
Although LwF works well on incremental classification and also anchor-based incremental object detection, we find that it shows no significant improvement compared to direct fine-tuning when applied to anchor-free object detectors.


In this paper, we revisit knowledge distillation and propose a Selective and Inter-related Distillation (SID) method for anchor-free fully convolutional detectors.
Instead of distilling from the source model outputs, we selectively perform distillation on non-regression outputs and some intermediate layers.
For different detectors, according to their unique network structures, performing distillation at the proper locations is critical for extracting noise-free source model knowledge.
By transferring the instance relations from the source model, our method preserves more old class knowledge in the target model.
In addition, to evaluate the capability of a detector under incremental conditions, a new evaluation metric is proposed.

\textbf{The contributions} of this paper are as follows:
\begin{itemize}
	\item To the best of our knowledge, we are the first to explore incremental detection on anchor-free fully convolutional object detectors.
	\item We propose a selective and inter-related distillation strategy for incremental anchor-free object detection.
	\item We propose a new evaluation metric, $\mathcal F1^i$ to better represent the performance of incremental detectors on both old and new classes.
	\item Using two different type anchor-free detectors, we demonstrate the superior performance of our proposed method on benchmark datasets under several incremental protocols.
\end{itemize}

\section{Related Work}
\label{Related Work}
In this section, we introduce the state-of-the-art anchor-free object detectors and then discuss incremental learning methods.

\subsection{Anchor-Free Object Detection}
\label{Anchor-free Object Detection}
YOLOv1 \cite{redmon2016you} is one of the very first works on anchor-free object detection.
It predicts the center of the objects and regresses the widths and heights of the respective bounding boxes.
However, to produce more precise detection, YOLOv1 only predicts bounding boxes at points near the center of objects which leads to low recall.
\citet{zhou2019objects} proposed CenterNet which regards each pixel within the feature map as a shape-agnostic anchor.
For each pixel within the feature map, CenterNet predicts whether it is an object center and regresses the width and height.
Due to its tiny structure, CenterNet does not need Non-Maximum Suppression (NMS) for post-processing and achieves a good speed-accuracy trade-off.
FCOS designed by \citet{tian2019fcos} is another recently proposed anchor-free detector.
For each object, FCOS predicts a point inside the object and regresses the distance from that point to the four sides of the bounding box.
Unlike YOLOv1 which only considers points near the center of objects, FCOS takes advantage of all points inside a ground truth box to predict the bounding box.
A multi-level Feature Pyramid Network (FPN) is used in FCOS to handle the low recall and ground truth box overlapping problems.
During post-processing, NMS is used to filter redundant bounding boxes.
FCOS achieves comparable detection accuracy to most anchor-based detectors.
On account of their model structures, we use FCOS (using FPN) and CenterNet (not using FPN) as our backbone networks to explore how different network structures will affect the anchor-free incremental detection.

\subsection{Incremental Learning}
\label{Incremental Learning}
Currently, one of the most popular incremental learning methods is to use extra regularization.
The regularization-based approaches can be further divided into prior-focused and data-focused.
Prior-focused methods penalize the important weights for old tasks to encourage them to stay unchanged.
The main difference between this type of methods is how they define the important weights for old tasks \citep{kirkpatrick2017overcoming, schwarz2018progress}.
Data-focused approaches use data from new task to approximate the performance of the previous tasks.
They penalize changes in the input-output function of the neural network for the old tasks.
\citet{li2017learning} first applied knowledge distillation \citep{hinton2015distilling} to incremental learning and proposed an incremental classifier called LwF.
During incremental learning, the new class data is passed to both the source model and the target model.
The target model is then trained using ground truth information for the new classes in addition to the source model output for the old classes.
The distillation loss will force the target model to follow the behavior of the source model on old tasks.

For this paper, we target on proposing a knowledge distillation based method for incremental object detection.
Compared to incremental classification, there are far fewer works on the more challenging problem of incremental object detection.
Using Fast RCNN as the backbone network, \citet{shmelkov2017incremental} adapted LwF to the detection task and proposed the first knowledge distillation based incremental detector.
Following \citet{shmelkov2017incremental}, some researchers have designed incremental learning methods on more advanced state-of-the-art object detectors.
\citet{Hao2019AnEA} proposed an incremental method based on Faster R-CNN \citep{ren2015faster}.
They divided the training data to multiple class groups to avoid missing annotations for old classes in the new data.
\citet{li2019rilod} and \citet{zhang2020class} designed incremental detectors based on RetinaNet \citep{lin2017focal}.
\citet{zhang2020class} proposed to handle the prediction bias between old and new classes using online unlabeled data.
\citet{perez2020incremental} designed a few-shot incremental detector based on CenterNet, but their method mainly focus on few-shot and meta learning.
Although much effort has been devoted to exploring incremental learning on detection tasks, to the best of our knowledge, there is no previous work exploring incremental learning for anchor-free detectors which are the latest research trend for object detection.

\begin{table*}[htpb]
	\centering
	\scriptsize{
		\caption{Ablation study of SID method based on FCOS \cite{tian2019fcos} using the VOC dataset under the one-step incremental protocol.}
		\begin{tabular}{|l|c|c|c|c|}
			\hline
			Note: a / b denotes overall mAP accuracy / $\mathcal F1^i$ score & + 1 class (20) & + 5 classes (16-20) & + 10 classes (11-20) & (1-20) \\
			\hline
			Directly fine-tuning on FCOS & 14.7\% / 20.5\% & 16.5\% / 9.7\% & 35.5\% / 10.3\% & \multirow{7}*{\tabincell{c}{71.6\% \\ (normal train)}}  \\
			\cline{1-4}
			LwF adapted to FCOS & 13.3\% / 18.6\% & 15.2\% / 9.6\% & 33.4\% / 11.5\% & \\
			\cline{1-4}
			Distill at `center-ness', `classification' & 36.0\% / 38.2\% & 26.1\% / 27.4\% & 38.6\% / 20.8\% & \\
			\cline{1-4}
			\tabincell{l}{Distill at `center-ness', `classification' \\ w `classification' replaced} & 42.0\% / 41.5\% & 37.0\% / 39.1\% & 41.5\% / 28.3\% & \\
			\cline{1-4}
			\tabincell{l}{Distill at `center-ness', `classification', CONV Tower \\ w `classification' replaced} & 68.1\% / \textbf{43.3\%} & 61.7\% / 48.1\% & 58.9\% / 58.8\% & \\
			\cline{1-4}
			\tabincell{l}{Distill at `center-ness', `classification', CONV Tower, \\ ResNet w `classification' replaced} & 68.1\% / 38.5\% & 61.1\% / 46.1\% & 58.5\% / 58.4\% & \\
			\cline{1-4}
			\tabincell{l}{Distill at `center-ness', `classification', CONV Tower \\ and inter-relation (2 samples) w `classification' replaced \\ (Our method)} & \textbf{68.3\%} / 42.4\% & \textbf{62.2\%} / \textbf{48.7\%} & \textbf{59.8\%} / \textbf{59.7\%} & \\
			\hline
		\end{tabular}	
		\label{tab:ablation on FCOS}
	}
\end{table*}

\section{Methodology}
\label{Methodology}
In this section, we propose a novel method as well as an evaluation metric for incremental object detection.

\subsection{Distillation at the Proper Locations}
Finding appropriate distillation locations is critical to achieve performance gain.
To deal with the different properties presented in different layers, we provide a detailed analysis of how to properly perform distillation.

\textbf{Output Distillation}:
For anchor-free fully convolutional object detectors, the regression outputs are partly trained only at positive pixels where new class objects exist.
The negative pixels which do not contain new class objects are not included in regression outputs training, so the model will output random predictions on background pixels.
Thus, the partially trained regression outputs are very noisy and cannot be directly used for distillation.
This has also been observed by \citet{zhang2020class}.
In their paper, \citet{zhang2020class} use an anchor-based fully convolutional object detector (RetinaNet) as their backbone.
They mention that for RetinaNet, the ratio of positive and negative anchor boxes is highly imbalanced and the negative boxes that correspond to background carry little information for knowledge distillation.
Thus, \citet{zhang2020class} propose an anchor box selection method to remove the noisy negative boxes.
Although the regression outputs are not suitable for distillation, regression prediction on old class objects is not significantly affected.
For negative pixels where new class objects do not exist and old class objects may indeed exist, the target model automatically follows the source model prediction since it is initialized by the source model parameters which are not retrained.
One the other hand, non-regression outputs, such as classification, are trained at all pixels, so there is little noise within these outputs.
Thus, for anchor-free fully convolutional object detectors, only the non-regression outputs will be used for distillation.

\textbf{Intermediate Distillation}:
The class-agnostic intermediate features are commonly used for distillation to preserve old class information.
However, we find that the efficiency of intermediate distillation strongly depends on the model structure.
If the detector model contains a FPN within its backbone, the intermediate distillation offers limited improvement.
The distillation at the final outputs which are connected with multiple feature maps from the FPN will back-propagate the front backbone layers and provide sufficient old model information.
On the other hand, if the detector model does not use the FPN, intermediate distillation can help to further alleviate catastrophic forgetting since distillation at the final output has limited long-range regularization effect on the front layers.
In our experiments, we use the $\mathcal L_2$ loss for both non-regression output and intermediate feature distillation.
The distillation loss is written as:

\begin{small}
	\begin{equation}
	\mathcal L_{Dist}=\sum \begin{Vmatrix} M_{t} - M_{s} \end{Vmatrix}_{2}^{2}
	\label{eq: dist}
	\end{equation}
\end{small}

\noindent where $t$ and $s$ refer to the target and the source networks, respectively.
$M$ is the feature map from intermediate layers or final outputs.
The distillation loss at different layers is added to form the final distillation loss.

\textbf{Restore Parameters on the Classification Layer}:
For a detection model, all parameters are shared between different classes except the final classification layer.
Although distillation has been performed to restrict the updating of important old class parameters, the target model still cannot obtain the same parameters of the source model which provide the best old class performance.
However, since the final classification layer is class-wise, the parameters for old class predictions are only related to old classes.
Thus, after target model training, we can replace the parameters of the target model by that of the source model at the final classification layer towards old categories.
Ablation studies in the Experiments Section corroborate our discoveries on output distillation, intermediate feature distillation, and the final classification layer.

\subsection{Inter-Related Distillation}
In conventional knowledge distillation based incremental learning methods, distillation is performed for each input image independently.
The inter-relation between features from different training instances is rarely considered.
However, the inter-relations between different instance features can help to provide extra source model information.
Inspired by \citet{liu2019knowledge} and \citet{park2019relational}, we transfer additional information from the source model by considering the relationship among different instances.
In our experiments, Euclidean distance is used to measure the inter-relations between instance features.
\begin{small}
	\begin{equation}
	\mathcal D_l(i, j)=\begin{Vmatrix} M_{l\_i} - M_{l\_j} \end{Vmatrix}_{2}^{2}, i, j = 1,...,I,
	\label{eq: IR distance}
	\end{equation}
\end{small}
where $D_l$ represents the feature distance in the $l^{th}$ layer, $M_{l\_i}$ and $M_{l\_j}$ refer to the $l^{th}$ layer feature map for image $i$ and $j$, respectively.
$I$ represents the total number of samples used for calculating inter-relations and the inter-relation is calculated in pair-wise between all $I$ samples.
The inter-related distillation loss is written as:
\begin{small}
	\begin{equation}
	\mathcal L_{IR\_Dist}=\sum_{L} \sum_{I} \begin{Vmatrix} D_{t\_l}(i, j) - D_{s\_l}(i, j) \end{Vmatrix}_{2}^{2}
	\label{eq: IR dist}
	\end{equation}
\end{small}
where $L$ represents the total number of layer outputs used for calculating the inter-relations.

\begin{figure}[tbp]
	\centering
	\includegraphics[width=8.6cm, keepaspectratio]{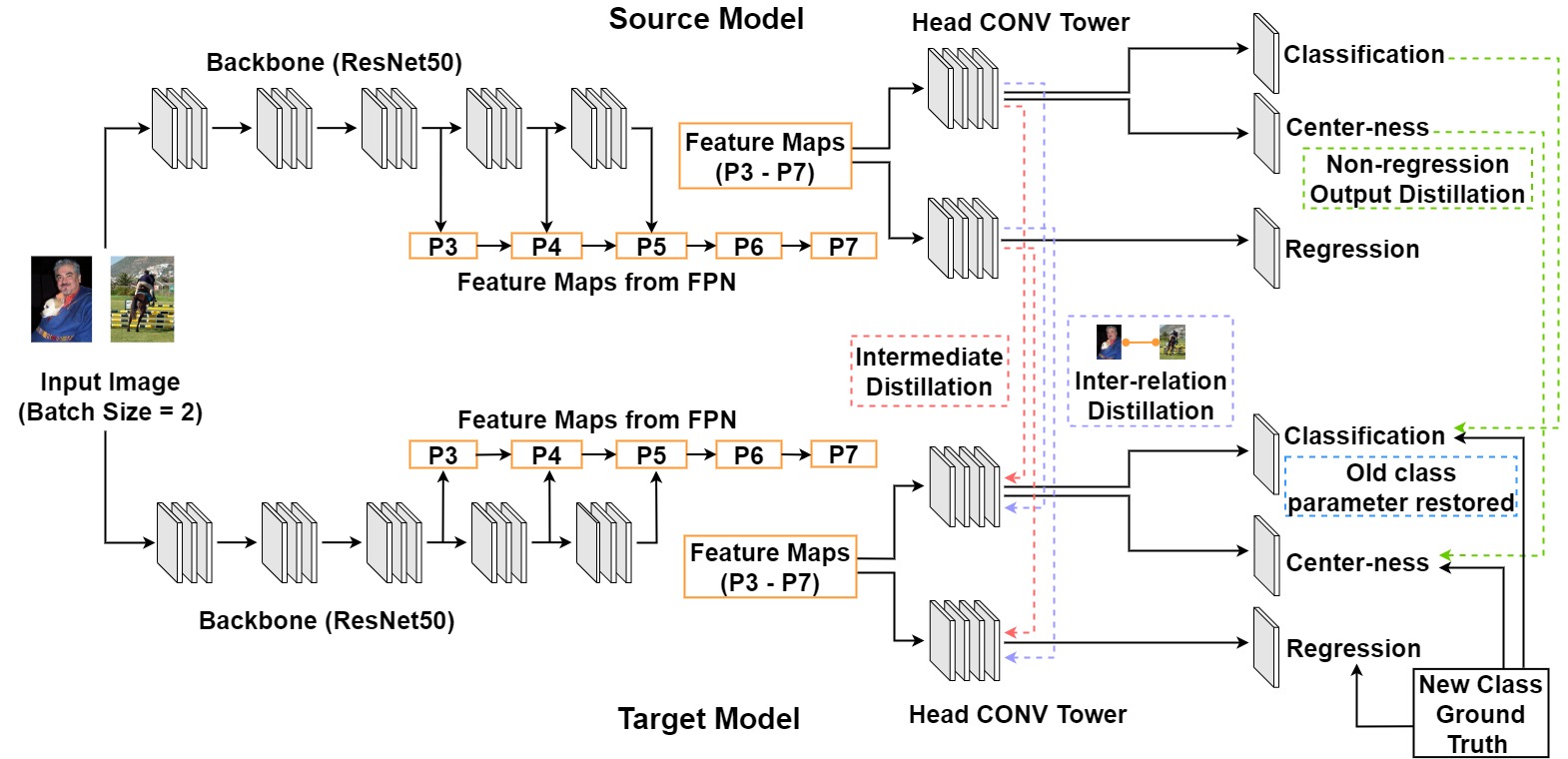}
	\caption{Framework of SID method based on FCOS \citep{tian2019fcos}.}
	\label{fig:SID based on FCOS}
\end{figure}

\subsection{Total Loss Function}
The overall loss ($\mathcal L_{total}$) is the weighted summation of the standard training loss, non-regression output and intermediate feature distillation loss \eqref{eq: dist}, and inter-related distillation loss \eqref{eq: IR dist}.
For standard training loss, we follow the loss function mentioned in the baseline model paper.
Hyper-parameters $\lambda_{1}$ and $\lambda_{2}$ help to balance each loss term and are empirically set to 1 in all our experiments.
\begin{small}
	\begin{equation}
	\mathcal{L}_{total}= \mathcal{L}_{Model} +\lambda_{1} \mathcal{L}_{Dist} +\lambda_{2} \mathcal{L}_{IR\_Dist}
	\label{eq: total Dist}
	\end{equation}
\end{small}

\subsection{Incremental Learning Score Metric ($\mathcal F1^i$)}
For conventional object detection tasks, the mean average precision (mAP) on all categories is commonly used for evaluating the performance.
However, this evaluation method is not appropriate for incremental learning since it does not differentiate between old and new class performance.
An ideal incremental detector needs to have the capability to preserve the old class performance with little affect on new class learning.
Inspired by the $\mathcal F1$ score, we propose a $\mathcal F1^i$ score which calculates the harmonic mean of the performance on the old and new classes.
It is written as:
\begin{small}
	\begin{equation}
	\mathcal F1^i = 2 \times \frac{P_o \times P_n}{P_o + P_n}
	\label{eq: incremental score}
	\end{equation}
\end{small}
where $P_o$ and $P_n$ represents the average mAP for old and new classes, respectively.
When the incremental detector performs equally well on both the old and new classes, the $\mathcal F1^i$ score is the same as the overall mAP.
When the incremental detector fails on either the old or new tasks, the $\mathcal F1^i$ score will be dominated by the smaller mAP, which indicates unbalanced behavior on incremental learning.
The commonly used overall mAP does not convey this important information.
Thus, it is preferable to use our proposed metric for incremental learning.

\section{Experiments}
\label{Experiments}
\begin{table*}[tpb]
	\centering
	\scriptsize{
		\caption{Ablation study of SID method based on CenterNet \cite{zhou2019objects} using the VOC dataset under the one-step incremental protocol.}
		\begin{tabular}{|l|c|c|c|c|}
			\hline
			Note: a / b denotes overall mAP accuracy / $\mathcal F1^i$ score & + 1 class (20) & + 5 classes (16-20) & + 10 classes (11-20) & (1-20) \\
			\hline
			Directly fine-tuning on CenterNet & 7.3\% / 9.0\% & 13.2\% / 5.0\% & 28.6\% / 0.0\% & \multirow{13}*{\tabincell{c}{65.6\% \\ (normal train)} } \\
			\cline{1-4}
			LwF adapted to CenterNet & 12.5\% / 10.9\% & 6.9\% / 1.5\% & 17.6\% / 0.0\% & \\
			\cline{1-4}
			Distill at `heatmap' layer-1 & 4.2\% / 2.9\% & 12.9\% / 2.6\% & 31.4\% / 0.0\% & \\
			\cline{1-4}
			Distill at `heatmap' layer-2 & 16.2\% / 22.3\% & 16.3\% / 9.2\% & 31.4\% / 0.0\% & \\
			\cline{1-4}
			Distill at `heatmap' layer-1 w `heatmap' layer-2 replaced & 32.0\% / 37.5\% & 24.8\% / 24.7\% & 33.4\% / 7.5\% & \\
			\cline{1-4}
			\tabincell{l}{Distill at `heatmap' layer-1, Up CONV \\ w `heatmap' layer-2 replaced} & 37.5\% / 40.8\% & 29.9\% / 31.9\%  & 33.8\% / 8.8\% & \\
			\cline{1-4}
			\tabincell{l}{Distill at `heatmap' layer-1, Up CONV, ResNet \\ w `heatmap' layer-2 replaced} & 40.6\% / \textbf{43.6\%} & 31.1\% / 33.3\% & 35.9\% / 15.5\% & \\
			\cline{1-4}
			\tabincell{l}{Distill at `heatmap' layer-1, Up CONV, ResNet \\ and inter-relation (2 samples) w `heatmap' layer-2 replaced} & \tabincell{l}{42.3\% $\pm$ 0.82\% \\ / 39.1\% $\pm$ 1.72\%} & \tabincell{l}{42.8\% $\pm$ 0.58\% \\ / 42.2\% $\pm$ 0.29\%} & \tabincell{l}{\textbf{49.1\% $\pm$ 0.38\%} \\ / \textbf{47.5\% $\pm$ 0.42\%}} & \\
			\cline{1-4}
			\tabincell{l}{Distill at `heatmap' layer-1, Up CONV, ResNet \\ and inter-relation (3 samples) w `heatmap' layer-2 replaced} & \tabincell{l}{43.7\% $\pm$ 0.38\% \\ / 36.4\% $\pm$ 1.48\%} & \tabincell{l}{49.7\% $\pm$ 0.16\% \\ / 44.7\% $\pm$ 0.25\%} & \tabincell{l}{46.2\% $\pm$ 0.18\% \\ / 46.2\% $\pm$ 0.17\%} & \\
			\cline{1-4}
			\tabincell{l}{Distill at `heatmap' layer-1, Up CONV, ResNet \\ and inter-relation (4 samples) w `heatmap' layer-2 replaced \\ (Our method)} & \tabincell{l}{\textbf{45.5\% $\pm$ 0.84\%} \\ / 27.0\% $\pm$ 2.35\%} & \tabincell{l}{\textbf{51.9\% $\pm$ 0.67\%} \\ / \textbf{46.1\% $\pm$ 1.00\%}} & \tabincell{l}{43.3\% $\pm$ 0.47\% \\ / 43.0\% $\pm$ 0.42\%} & \\
			\hline
		\end{tabular}	
		\label{tab:ablation on CenterNet}
	}
\end{table*}
In this section, to prove the effectiveness of each part of our proposed method, we first perform ablation studies on the VOC dataset using two baseline detectors --- FCOS and CenterNet.
Then we perform experiments on several one-step and multi-step incremental scenarios on both the VOC and COCO datasets to validate our method.

\subsection{Model Implementation Details}
We build our method on top of FCOS \citep{tian2019fcos} and CenterNet \citep{zhou2019objects} using their public implementations.
Both models are implemented using Pytorch framework.
For FCOS, ResNet-50 \citep{he2016deep} and FPN \citep{lin2017focal} are used as its backbone.
We follow the training strategy of \citet{shmelkov2017incremental} to train our model.
For the first training step, the network is trained by 40k iterations for VOC and 400k iterations for COCO.
In the following incremental steps, the network is trained for 10k iterations when only one class is added and the same number of iterations as the first step if multiple classes are added at once.
The learning rate is set to 0.001, decaying to 0.0001 after 30k iterations.
For CenterNet \citep{zhou2019objects}, ResNet-50 is used as its backbone.
We follow the same training strategy mentioned in their paper to train our model.
The network is trained by 70 epochs for each incremental step for both the VOC and COCO datasets.
The learning rate is set to 0.000125, decaying by 10 times at 45 and 60 epochs.

\subsection{Dataset and Evaluation Metric}
Two detection benchmark datasets are used, PASCAL VOC 2007 \citep{everingham2010pascal} and COCO 2014 \citep{lin2014microsoft}.
For experiments on VOC, we use train and validation set for training and test set for testing.
For experiments on COCO, we use train and valminusminival set for training and minival set for testing.
For the evaluation metric, we use mAP at 0.5 Intersection over Union (IoU) for both datasets and also use mAP weighted across various IoUs from 0.5 to 0.95 for COCO.
In addition, the proposed $\mathcal F1^i$ score is calculated to evaluate the balance between old and new class performance.

Incremental object detection requires several important settings for the training data.
First, we need to specify the sequence of the new categories being provided to the model.
We follow the widely used protocol in current literature \citep{shmelkov2017incremental, zhang2020class} --- the new categories are continually added in alphabetical order.
Second, we specify the accessible data for each incremental step.
Storing representative old data exemplars \citep{rebuffi2017icarl, castro2018end} or using unlabeled online data \citep{zhang2020class} are commonly used in many incremental learning methods to boost the model performance.
In our experiments, no old data or extra online data is provided.
In each incremental training step, only the training data for the new classes is available.
Third, we specify how to handle old class objects appearing in the new class data.
For object detection, each input image may contain multiple classes of objects.
In incremental detection, the object classes can come from both old and new categories.
Similar to the work in \citet{shmelkov2017incremental}, to make the annotating procedure much closer to real-life conditions, the labels for old class objects are not provided in new class data.

\begin{figure}[tbp]
	\centering
	\includegraphics[width=8.5cm, keepaspectratio]{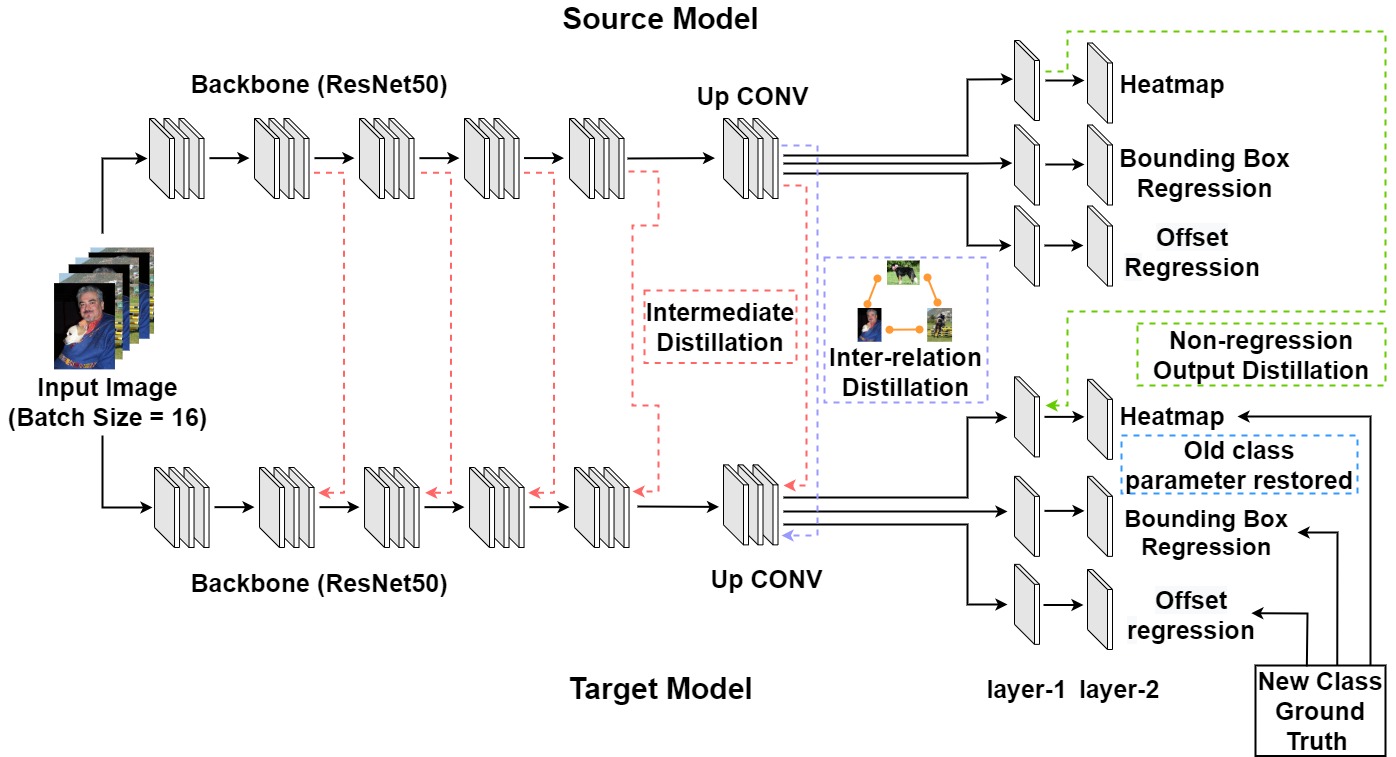}
	\caption{Framework of SID method based on CenterNet \citep{zhou2019objects}.}
	\label{fig:SID based on CenterNet}
\end{figure}

\begin{table*}[tpb]
	\centering
	\scriptsize{
		\caption{Performance for SID based on FCOS under \emph{the three-step incremental adding five new classes at a time} protocol on the VOC dataset.
			A(a-b) is the first-step normal training for categories a to b and +B(c-d) is the incremental learning for categories c to d.}
		\begin{tabular}{|l|c|c|c|c|c|}
			\hline
			Note: a / b denotes overall mAP accuracy / $\mathcal F1^i$ score & A(1-5) & +B(6-10) & +B(11-15) & +B(16-20) & A(1-20) \\
			\hline
			Directly fine-tuning (Catastrophic forgetting) & \multirow{4}*{70.6\%} & 37.2\% / 8.2\% & 23.0\% / 0.7\% & 12.4\% / 1.4\% &  \multirow{4}*{71.6\%} \\
			\cline{1-1} \cline{3-5}
			EWC \citep{kirkpatrick2017overcoming} &  & 44.1\% / 33.8\% & 28.5\% / 16.5\% & 18.3\% / 16.0\% & \\
			\cline{1-1} \cline{3-5}
			Online EWC \citep{schwarz2018progress} &  & 44.1\% / 33.8\% & 29.3\% / 18.7\% & 18.7\% / 16.7\% & \\
			\cline{1-1} \cline{3-5}
			SID (Our method) &  & \textbf{58.8\%} / \textbf{58.6\%} & \textbf{44.9\%} / \textbf{46.1\%}& \textbf{36.2\%} / \textbf{31.9\%} &  \\
			\hline
		\end{tabular}	
		\label{tab: 5_5_5_5_FCOS}
	}
	\centering
	\scriptsize{
		\caption{Performance for SID based on FCOS under \emph{the five-step incremental adding two new classes at a time} protocol on the VOC dataset.}
		\begin{tabular}{|l|c|c|c|c|c|c|c|}
			\hline
			\tabincell{l}{Note: a / b denotes \\ overall mAP accuracy \\ / $\mathcal F1^i$ score} & A(1-10) & +B(11-12) & +B(13-14) & +B(15-16) & +B(17-18) & +B(19-20) & A(1-20) \\
			\hline
			\tabincell{l}{Directly fine-tuning} & \multirow{2}*{73.4\%} & 14.9\% / 14.8\% & 7.8\% / 2.9\% & 7.8\% / 0.0\% & 4.1\% / 1.1\% & 6.1\% / 2.7\% & \multirow{2}*{71.6\%} \\
			\cline{1-1} \cline{3-7}
			\tabincell{l}{SID (Our method)} & & \textbf{62.7\%} / \textbf{53.7\%} & \textbf{54.1\%} / \textbf{43.1\%} & \textbf{48.6\%} / \textbf{49.4\%} & \textbf{43.8\%} / \textbf{29.0\%} & \textbf{39.8\%} / \textbf{33.0\%} & \\
			\hline
		\end{tabular}	
		\label{tab: 10_2_2_2_2_2_FCOS}
	}
	\centering
	\scriptsize{
		\caption{Performance for SID based on FCOS under \emph{the five-step incremental adding one new classes at a time} protocol on the VOC dataset.}
		\begin{tabular}{|l|c|c|c|c|c|c|c|}
			\hline
			\tabincell{l}{Note: a / b denotes \\ overall mAP accuracy \\ / $\mathcal F1^i$ score} & A(1-15) & +B(16) & +B(17) & +B(18) & +B (19) & +B(20) & A(1-20)\\
			\hline
			\tabincell{l}{Directly fine-tuning} & \multirow{2}*{73.71\%} & 21.1\% / 26.0\% & 10.3\% / 12.3\% & 4.8\% / 6.5\% & 1.4\% / 0.5\% & 2.5\% / 0.0\% & \multirow{2}*{71.6\%} \\
			\cline{1-1} \cline{3-7}
			\tabincell{l}{SID (Our method)} & & \textbf{68.0\%} / \textbf{31.5\%} & \textbf{63.0\%} / \textbf{31.5\%} & \textbf{57.3\%} / \textbf{29.0\%} & \textbf{53.2\%} / \textbf{32.4\%} & \textbf{48.9\%} / \textbf{35.9\%} & \\
			\hline
		\end{tabular}	
		\label{tab: 15_1_1_1_1_1_FCOS}
	}
\end{table*}

\subsection{Ablation Study}
To validate the effectiveness of each part of our proposed method, we perform experiments on three one-step incremental learning scenarios on the VOC dataset.
Table \ref{tab:ablation on FCOS} and \ref{tab:ablation on CenterNet} show the experimental results based on FCOS and CenterNet, respectively.
When directly fine-tuning the model on new class data, catastrophic forgetting happens and both models' performance dramatically drops.
Both tables show that LwF which works well for incremental classification does not provide better performance than directly fine-tuning.
In addition, the performance of incremental detection strongly depends on the baseline model.
It is quite misleading to directly compare different incremental detectors based on fundamentally different baseline models.
\citet{shmelkov2017incremental}, \citet{li2019rilod} and \citet{Hao2019AnEA} perform distillation at the model output and inter-mediate features based on Fast RCNN, RetinaNet, and Faster RCNN, respectively.
In our ablation study, we extensively explored the effect of final output distillation at different output locations, such as classification and regression outputs, and feature distillation at different intermediate layers.
Part of our ablation study can be regarded as comparisons with these works but reproduced on our anchor-free backbone models.

Figure \ref{fig:SID based on FCOS} shows the framework of our SID method on FCOS \citep{tian2019fcos}.
The model output of FCOS comprises three parts: `classification', `center-ness' and `regression'.
It predicts each object as a point and a 4D vector containing the distance from that point to the four sides of the bounding box.
As the `regression' branch is partially trained only at positive pixels, the `regression' branch will output random numbers at locations without objects.
Therefore, the raw outputs from FCOS, especially the `regression' output, contain correct predictions as well as strong noise components.
Compared with LwF which distills at all model outputs, only distilling at non-regression outputs (`center-ness' and `classification') provides 22.7\%, 10.9\% and 5.2\% (19.6\%, 17.8\% and 9.3\%) mAP ($\mathcal F1^i$) improvement on each incremental scenario for adding 1, 5 and 10 new classes, respectively.
Then by simply replacing the old class parameters of the final classification layer, the distilled model has 6.0\%, 10.9\% and 2.9\% (3.3\%, 11.7\% and 7.5\%) mAP ($\mathcal F1^i$) improvement on each scenario.
After that, an intermediate distillation is applied at the 4-layer Head CONV Tower outputs which provide 26.1\%, 24.7\% and 17.4\% (1.8\%, 9.0\% and 30.5\%) mAP ($\mathcal F1^i$) improvement on each scenario.
Although the intermediate distillation at Head CONV Tower outputs significantly alleviates catastrophic forgetting, the intermediate distillation at ResNet outputs is not effective.
FPN is used in FCOS structure, so distillation at non-regression and Head CONV Tower outputs will be back-propagated to feature maps from different backbone layers.
Further distillation at the ResNet layers cannot provide extra information.
The inter-related distillation is also applied at the Head CONV Tower outputs and is calculated between 2 training samples.
It helps to provide a further overall accuracy improvement.

Figure \ref{fig:SID based on CenterNet} shows the framework of our SID method on CenterNet \citep{zhou2019objects}.
The output of CenterNet comprises three parts: `heatmap', `regression' and `offset'.
It predicts each object as a center point and two 2D vectors.
One vector (`regression') represents the width and height of the bounding box.
Another (`offset') represents the offsets of center point location towards the horizontal and vertical axis.
Similar to FCOS, the raw outputs from  CenterNet, especially the `regression' and `offset' outputs, contain strong noise.
Different from FCOS which only has one convolutional layer for each output branch, CenterNet consists of two convolutional layers for each output branch.
The first convolutional layer is class-agnostic, and the second convolutional layer is class-wise.
We find that for the non-regression output branch (`heatmap'), distilling at the first convolutional layer output and replacing the old class parameters for the second convolutional layer provides the best results.
Compared to LwF, this selective distillation operation shows 19.5\%, 17.9\% and 15.8\% (26.6\%, 23.2\% and 7.5\%) mAP ($\mathcal F1^i$) improvement on each incremental scenario in Table \ref{tab:ablation on CenterNet}.

\begin{table*}[tpb]	
	\centering
	\scriptsize{
		\caption{
			Performance for one-step incremental experiments on the COCO dataset.
		}
		\begin{tabular}{|l|c|c|c|c|}
			\hline
			\tabincell{l}{Note: a / b denotes \\ overall mAP accuracy / $\mathcal F1^i$ score} & + 5 classes (76-80) & + 10 classes (71-80) & + 40 classes (41-80) & normal train (1-80) \\
			\hline
			\tabincell{l}{Directly fine-tuning (mAP@.5)} & 3.3\% / 0.0\% & 6.4\% / 0.0\% & 25.4\% / 0.5\% & \multirow{2}*{47.0\%} \\
			\cline{1-4}
			SID (Our method) (mAP@.5) & \textbf{46.8\%} / \textbf{48.4\%} & \textbf{46.4\%} / \textbf{46.6\%} & \textbf{41.6\%} / \textbf{41.2\%} & \\
			\hline
			\tabincell{l}{Directly fine-tuning (mAP@[.5, .95])} & 2.2\% / 0.0\% & 4.1\% / 0.0\% & 16.2\% / 0.3\% & \multirow{2}*{29.3\%} \\
			\cline{1-4}
			SID (Our method) (mAP@[.5, .95]) & \textbf{28.9\%} / \textbf{29.9\%} & \textbf{28.8\%} / \textbf{28.7\%} & \textbf{25.2\%} /  \textbf{24.9\%} & \\
			\hline
		\end{tabular}
		\label{tab:coco_one_step}}
	\centering
	\scriptsize{
		\caption{
			Performance for multi-step incremental experiments on the COCO dataset.
			A(a-b) is the first-step normal training for categories a to b and +B(c) is the incremental learning for category c.}
		\begin{tabular}{|l|c|c|c|c|c|c|c|}
			\hline
			\tabincell{l}{Note: a / b denotes \\ overall mAP accuracy \\ / $\mathcal F1^i$ score} & A(1-75) & +B(76) & +B(77) & +B(78) & +B(79) & +B(80) & A(1-80) \\
			\hline
			\tabincell{l}{Directly fine-tuning \\ (mAP@.5)} & \multirow{2}*{46.8\%} & 0.6\% / 0.0\% & 0.4\% / 0.0\% & 0.3\% / 0.0\% & 0.3\% / 0.0\% & 0.5\% / 0.0\% & \multirow{2}*{47.0\%} \\
			\cline{1-1} \cline{3-7}
			\tabincell{l}{SID (Our method) \\ (mAP@.5)} & & \textbf{43.5\%} / \textbf{46.3\%} & \textbf{40.2\%} / \textbf{32.8\%} & \textbf{35.8\%} / \textbf{29.7\%} & \textbf{33.2\%} / \textbf{34.3\%} & \textbf{31.1\%}  / \textbf{42.3\%} & \\
			\hline
			\tabincell{l}{Directly fine-tuning \\ (mAP@[.5, .95])} & \multirow{2}*{29.4\%} & 0.5\% / 0.0\% & 0.3\% / 0.0\% & 0.2\% / 0.0\% & 0.2\% / 0.0\% & 0.4\% / 0.0\% & \multirow{2}*{29.3\%} \\
			\cline{1-1} \cline{3-7}
			\tabincell{l}{SID (Our method) \\ (mAP@[.5, .95])} & & \textbf{26.4\%} / \textbf{29.7\%} & \textbf{24.3\%} / \textbf{19.5\%} & \textbf{21.4\%} / \textbf{18.2\%} & \textbf{19.8\%}  / \textbf{20.5\%} & \textbf{18.5\%}  / \textbf{26.1\%} & \\
			\hline
		\end{tabular}
		\label{tab:coco_multi_step}}
\end{table*}

After that, intermediate distillation is applied at Up CONV and ResNet outputs which provide 5.5\%, 5.1\%, 0.4\% (3.3\%, 7.2\%, 1.3\%) and 3.1\%, 1.2\%, 2.1\% (2.8\%, 1.4\%, 6.7\%) mAP ($\mathcal F1^i$) gain on each one-step incremental scenario.
On the contrary to FCOS, as in CenterNet FPN is not used, distillation at ResNet outputs helps to extract model information from front layers and shows improvement.
During the incremental training for CenterNet, the batch size is set to 16.
Considering computational expense, we choose 2, 3 or 4 samples to calculate their inter-relations.
The samples are randomly picked within the batch for each iteration.
As there is randomness in sampling, we perform each of the experiment three times and calculate the mean and deviation.
For one-step adding 1 new class scenario, the best mAP is obtained when 4 samples are used and the improvement is 4.9\%.
In this setting, although using inter-related distillation improves the overall accuracy, the $\mathcal F1^i$ score decreases.
Under the same incremental scenario, similar result is also obtained on FCOS based model.
For incremental scenarios that the old and new class ratio is large (19:1 for this case), the retrained model suffers from more severe catastrophic forgetting problem.
In such settings, a stronger distillation can help to maintain more old class knowledge but inevitably hurting the performance of new classes.
As in this setting, the only one new class absorbing all the negative effects, which causes large accuracy difference between old and new classes, and leads to a lower $\mathcal F1^i$ score.
For one-step adding 5 new classes scenario, the best performance is obtained when 4 samples are used and the mAP ($\mathcal F1^i$) improvement is 20.8\% (12.8\%).
For one-step adding 10 new classes scenario, the best performance is obtained when 2 samples are used and the mAP ($\mathcal F1^i$) improvement is 13.2\% (32.0\%).

Under different incremental settings, the best number of samples for inter-related distillation is varied.
We think this is because, for different incremental settings, the extent of catastrophic forgetting is different due to the amount of training data, the ratio between old and new classes, and the percentage of old class missing annotations.
In addition, the improvement from inter-related distillation for CenterNet based model is much higher than FCOS based model.
We conjecture the training batch size is the reason.
A larger batch size will help to provide more sufficient inter-relation information.
According to their default implementation, the batch size is set to 16 for CenterNet and 2 for FCOS.

\subsection{Experiments on the VOC Dataset}
Using FCOS as the backbone network, we have investigated the results under multi-step incremental scenarios for the VOC dataset.
Observing from Table \ref{tab: 5_5_5_5_FCOS}, \ref{tab: 10_2_2_2_2_2_FCOS} and \ref{tab: 15_1_1_1_1_1_FCOS}, our SID method outperforms directly fine-tuning for each incremental step on all three different multi-step settings.
The average mAP ($\mathcal F1^i$) improvement is 22.4\%, 41.7\% and 50.1\% (42.1\%, 37.3\% and 23.0\%) for scenarios on Table \ref{tab: 5_5_5_5_FCOS}, \ref{tab: 10_2_2_2_2_2_FCOS} and \ref{tab: 15_1_1_1_1_1_FCOS}, respectively.
We find that SID achieves nearly the same overall mAP and $\mathcal F1^i$ score in some cases such as in (+B(6-10)) step in Table \ref{tab: 5_5_5_5_FCOS}, which indicates the well-balanced performance of our detector.
The gap between $\mathcal F1^i$ score and mAP shows the degree of the unbalanced learning in different scenarios.
We conjecture that the gap is due to the different difficulty levels of different classes, as well as the unbalanced numbers of the old and new tasks.
On the contrast, the $\mathcal F1^i$ score of fine-tuning is far lower than our method, since the old class mAP of fine-tuning is nearly zero and denominates the $\mathcal F1^i$ score.
As the performance of LwF degrades to 0 in the second incremental step, we do not put these results in the table due to the space limitation.
In addition, under three-step incremental adding five new classes scenario, we also compared our method with EWC \citep{kirkpatrick2017overcoming} and online EWC \citep{schwarz2018progress}.
The average mAP ($\mathcal F1^i$) improvement is 16.3\% and 15.9\% (23.4\% and 22.5\%) comparing with EWC and online EWC, respectively.

\subsection{Experiments on the COCO Dataset}
Using FCOS as the backbone network, we have also performed experiments on the COCO dataset under both one-step and multi-step incremental scenarios.
According to Table \ref{tab:coco_one_step}, under three different one-step scenarios, our SID method outperforms directly fine-tuning by a large margin on both average mAP and $\mathcal F1^i$ score.
Table \ref{tab:coco_multi_step} shows the experimental results under five-step incremental of each time adding one new class protocol.
The overall average gain is 36.4\% mAP at 0.5 IoU and 21.7\% mAP at weighted across different IoU from 0.5 to 0.95.
The overall $\mathcal F1^i$ score gain is 37.1\% at 0.5 IoU and 22.8\% at weighted across different IoU from 0.5 to 0.95.

\section{Conclusion}
In this paper, we focus on designing an incremental paradigm for anchor-free fully convolutional object detectors.
Based on knowledge distillation, a novel incremental detection method SID is proposed.
First, conventional distillation is applied at non-regression model outputs and suitable intermediate layer outputs.
Second, inter-related distillation is applied to provide extra high order source model information and further improve the distillation quality.
Finally, after training, the target model's old class parameters on the class-wise final classification layer are replaced with the corresponding parameters from the source model to further alleviate catastrophic forgetting.
In addition, to better evaluate the balance performance between old and new classes, a new metric is proposed.
Experiments on benchmark datasets demonstrate the effectiveness of our method.

\section*{Acknowledgments}
This research was funded by the Australian Government through the Australian Research Council and Sullivan Nicolaides Pathology under Linkage Project LP160101797.

\bibliographystyle{model2-names}
\bibliography{refs}

\end{document}